\documentclass{article}

\newcommand{\etal}{\textit{et al}.}
\newcommand{\ie}{\textit{i}.\textit{e}.}

\usepackage[numbers]{natbib}

\usepackage[final]{neurips_2022}

\usepackage[utf8]{inputenc} 
\usepackage[T1]{fontenc}    
\usepackage{hyperref}       
\usepackage{url}            
\usepackage{booktabs}       
\usepackage{amsfonts}       
\usepackage{nicefrac}       
\usepackage{microtype}      
\usepackage{xcolor, eucal}         
\usepackage{array}

\usepackage{amsmath} 
\usepackage{graphicx}  
\usepackage{multirow} 
\usepackage[labelfont=bf,font=small]{caption} 

\usepackage{graphicx}
\usepackage{floatrow}

\usepackage{subcaption} 

\usepackage{bm}
\usepackage{makecell} 

\newcolumntype{C}{ >{\centering\arraybackslash} m{4.5cm} }
\newcolumntype{D}{ >{\centering\arraybackslash} m{2.0cm} }

\newcolumntype{E}{ >{\centering\arraybackslash} m{10.0cm} }
\newcolumntype{F}{ >{\centering\arraybackslash} m{1.0cm} }

\newfloatcommand{figurebox}{figure}[\nocapbeside][\dimexpr(\textwidth-\columnsep)/2\relax]
\newfloatcommand{tablebox}{table}[\nocapbeside][\dimexpr(\textwidth-\columnsep)/2\relax]

\def\SP{~~~~~~}

\title{StyleAvatar3D: Leveraging Image-Text Diffusion Models for High-Fidelity 3D Avatar Generation}

\author{
Chi Zhang$^1$,
\SP 
 Yiwen Chen$^2$,
\SP 
Yijun Fu$^1$,
\SP 
Zhenglin Zhou$^1$,
\SP 
Gang Yu$^1$\thanks{Corresponding author \\Project page: \url{https://github.com/icoz69/StyleAvatar3D}}, \\
\And
\SP 
Zhibin Wang$^1$,
\SP 
Bin Fu$^1$,
\SP
Tao Chen$^3$,
\SP 
Guosheng Lin$^2$,
\SP 
Chunhua Shen$^4$
\\[0.1325cm]
$ ^1$Tencent PCG, China
\SP 
$ ^2$Nanyang Technological University, Singapore
\SP 
$ ^3$Fudan University, China\\
\SP 
$ ^4$Zhejiang University, China
}

\begin{document}

\makeatletter
\let\@oldmaketitle\@maketitle%
\renewcommand{\@maketitle}{\@oldmaketitle%
 \centering
    \includegraphics[width=1\textwidth]{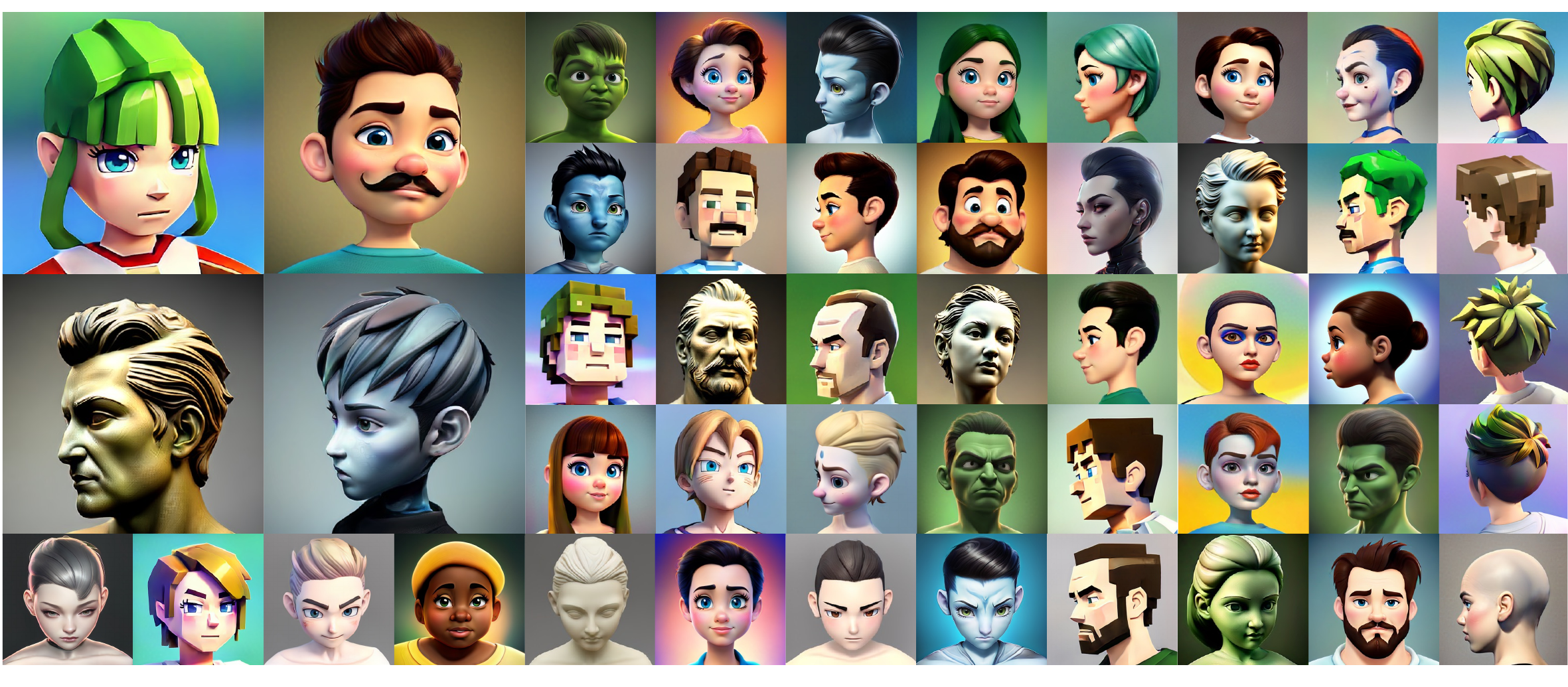}
    \vskip -0.5em
     \captionof{figure}{ 
     Visualization of 3D stylized avatars produced by our  framework. The styles of avatars  can be defined using  text prompts or example images.
     }
    \label{fig:teaser}
    \bigskip}                   %
\makeatother

\maketitle

\begin{abstract}
The recent advancements in image-text diffusion models have stimulated research interest in large-scale 3D generative models. Nevertheless, the limited availability of diverse 3D resources presents significant challenges to learning. 
In this paper, we present a novel method for generating high-quality, stylized 3D avatars that utilizes pre-trained image-text diffusion models for data generation and a Generative Adversarial Network (GAN)-based 3D generation network for training.
Our method leverages the comprehensive priors of appearance and geometry offered by image-text diffusion models to generate multi-view images of avatars in various styles. 
During data generation, we employ poses extracted from existing 3D models to guide the generation of multi-view images. 
To address the misalignment between poses and images in  data, we investigate view-specific prompts and develop a coarse-to-fine discriminator for GAN training. 
We also delve into attribute-related prompts to increase the diversity of the generated avatars.
Additionally, we develop a latent diffusion model within the style space of StyleGAN to enable the generation of avatars based on image inputs. Our approach demonstrates superior performance over current state-of-the-art methods in terms of visual quality and diversity of the produced avatars.

\end{abstract}

\section{Introduction}

In recent years, generative models have made significant strides in generating high-fidelity 2D images~\cite{nichol2021glide, ramesh2022hierarchical, stable_diffusion, imagen_saharia2022photorealistic}, primarily due to the availability of large-scale image-text pairs~\cite{radford2021learning} and advanced generative model architectures such as diffusion models~\cite{song2020score, ho2020denoising, dhariwal2021diffusion, ho2022classifier, yang2022diffusion}. These models enable users to generate realistic images using text prompts, obviating the need for manual intervention.
However, 3D generative models~\cite{3dgan, gao2022get3d, luo2021diffusion, zhou20213d, zeng2022lion} still face considerable challenges owing to the limited availability and diversity of 3D models for learning, compared to their 2D counterparts. 
The manual creation of 3D assets in software engines is a laborious process that demands significant expertise, limiting the availability of diverse and high-quality 3D models~\cite{guo2019relightables, lombardi2018deep, wood2021fake}. To tackle this issue, researchers~\cite{nichol2022point, jain2022zero, poole2022dreamfusion, lin2023magic3d} have recently explored pre-trained image-text generative models for generating high-fidelity 3D models. These models provide rich priors of object appearance and geometries, which can potentially facilitate the generation of realistic and diverse 3D models.

In this paper, we propose a novel approach to 3D stylized avatar generation that leverages pre-trained text-to-image diffusion models, enabling users to define styles and facial attributes of avatars using text prompts. Specifically, we employ EG3D~\cite{Chan2022eg3d}, a GAN-based 3D generation network, as our generator, which offers several advantages. Firstly, EG3D~\cite{Chan2022eg3d} requires calibrated images for training instead of 3D data, allowing for continuous improvement of the diversity and fidelity of 3D models through enhanced image data, a relatively straightforward task for 2D images. Secondly, since the images used for training do not necessitate strict multi-view consistency in appearance, we can generate each view separately, effectively managing the randomness during image generation. To generate calibrated 2D training images for training EG3D~\cite{Chan2022eg3d}, our approach utilizes ControlNet~\cite{zhang2023adding} built upon StableDiffusion~\cite{stable_diffusion}, which enables image generation guided by predefined poses. These poses can be either synthesized or extracted from avatars in existing engines, allowing for the reuse of camera parameters from pose images for learning purposes.

ControlNet~\cite{zhang2023adding} often struggles to generate views with large angles, such as the backside of the head, even when using correct pose images as guidance. These failure outputs pose significant challenges for generating full 3D models. To address this issue, we have approached the problem from two different perspectives. Firstly, we have designed view-specific prompts for different views during image generation, significantly reducing the number of failure cases. Nevertheless, even with view-specific prompts, the synthesized images may not be perfectly aligned with the pose images. To tackle this misalignment, we have developed a coarse-to-fine discriminator for  training 3D GANs. In our framework, each image data is associated with a coarse pose annotation and a fine pose annotation. During GAN training, we randomly choose an annotation for training. For confident views such as the front face, we assign a high probability of using the fine pose annotation, while the learning of the rest views relies more on coarse views. This strategy enables us to generate more accurate and diverse 3D models, even when the input images have noisy annotations.

To enable conditional 3D generation with an image input, we have also developed a latent diffusion model~\cite{ramesh2022hierarchical} in the latent style space of StyleGAN~\cite{karras2019style, karras2020analyzing}. The style code is low-dimensional, highly expressive, and compact, making the diffusion model fast to train. We directly sample image and style code pairs from our trained 3D generators as training data to learn the diffusion model~\cite{ramesh2022hierarchical}.

To evaluate the effectiveness of our proposed method, we conducted extensive experiments on various large-scale datasets. Our results demonstrate that our method outperforms existing state-of-the-art methods in terms of both visual quality and diversity. In summary, this paper presents a novel framework for generating high-fidelity 3D avatars that leverages pre-trained image-text diffusion models. Our framework allows styles and face attributes to be defined by text prompts, significantly enhancing the flexibility of avatar generation. Additionally, we have proposed a coarse-to-fine pose-aware discriminator to address the image-pose misalignment issue, enabling better utilization of image data with inaccurate pose annotations. Lastly, we have developed an extra conditional generation module that allows for conditional 3D generation with image input in the latent style space, further increasing the flexibility of our framework and allowing users to generate 3D models tailored to their individual preferences.

\section{Related Work}

This section provides an overview of related literature in the field of 3D generation from 2D images.

\textbf{Text-to-image Generative Models.}
Tremendous advancements have been made in the field of text-to-image generative tasks~\cite{song2020denoising, nichol2021glide, preechakul2021diffusion, mou2023t2i}. Several models, such as StableDiffusion~\cite{stable_diffusion}, Imagen~\cite{imagen_saharia2022photorealistic}, and DALL-E 2~\cite{ramesh2022hierarchical}, have been proposed to generate images guided by encoded text prompts. These models integrate text guidance during the diffusion process in a classifier-free manner, allowing for the customization of image styles, contents, and details through prompt adjustment. 
StableDiffusion~\cite{stable_diffusion} is particularly notable among these methods because it carries out the diffusion process in the latent space of an autoencoder, leading to reduced inference speed and memory cost. ControlNet~\cite{zhang2023adding} extends StableDiffusion by introducing additional parametric modules to a pre-trained model, providing accurate control over the output image content. ControlNet can support various types of guidance and allows the use of multiple guidance at the same time, such as depth images~\cite{ranftl2020towards,zhang2022hierarchical}, human poses~\cite{openpose, kreiss2021openpifpaf}, edge maps~\cite{canny1986computational}, and others. 
Low-Rank Adaptation (LoRA)~\cite{hu2021lora}, originally used for finetuning large language models, is recently introduced for finetuning StableDiffusion, which can generate images of a subject given only a few training images of it.

\textbf{3D generation based on pre-trained image generative models.}
Directly transferring the success of image diffusion models to 3D is a challenge, as significant amounts of 3D data with appropriate representations must be prepared for learning. Wang~\etal~\cite{wang2022rodin} develop a 3D digital avatar diffusion model called Rodin, which represents 3D data using parametric tri-plane~\cite{Chan2022eg3d} features obtained by fitting existing 3D avatar models in the engine.
Image generative models possess significant capabilities and rich priors, motivating researchers to explore their potential for 3D generation.
For instance, Dreamfusion~\cite{poole2022dreamfusion} designs a Score Distillation Sampling method to extract knowledge from image generative models to optimize Neural Radiance Fields (NeRFs)~\cite{mildenhall2020nerf, barron2022mip} guided by prompts. However, this model generation relies on slow optimization of NeRFs, and the diversity of the models remains uncertain.
Magic3D~\cite{lin2023magic3d} improves the learning speed and the resolution of DreamFusion~\cite{poole2022dreamfusion} by adopting a two-stage optimization strategy that utilizes a sparse 3D hash grid structure.
DreamBooth3D~\cite{raj2023dreambooth3d} combines DreamBooth~\cite{ruiz2022dreambooth} and DreamFusion~\cite{poole2022dreamfusion}, which allows personalizing text-to-3D generative models from a few images of a subject.
Latent-NeRF~\cite{metzer2022latent} improves DreamFusion~\cite{poole2022dreamfusion} by applying the diffusion process in a latent space of a pre-trained autoencoder.

\textbf{Domain Adaptation of 3D GANs.}
Our work is closely related to EG3D~\cite{Chan2022eg3d}, a geometry-aware GAN based on StyleGAN~\cite{karras2019style, karras2020analyzing}. EG3D's generator produces features of three orthogonal 2D planes, and it uses volume rendering to generate various views of the 3D model by sampling points from these planes. In contrast to the raw StyleGAN~\cite{karras2019style, karras2020analyzing} discriminator, EG3D's pose-conditioned discriminator requires accurate camera pose annotations of images to learn multi-view consistent 3D models.
For real-face datasets like FFHQ~\cite{karras2019style}, pose estimators~\cite{openpose,zhang2014facial,bulat2017far,huang2021adnet} are commonly used to obtain pose annotations. However, these estimators may fail to detect faces at large angles, resulting in incomplete 3D models that are 2.5D rather than full 3D~\cite{bulat2017far}. Additionally, the use of such detectors may be limited for stylized images, which reduces the applicability of the method~\cite{yaniv2019face}.
PoF3D~\cite{shi2023learning} develops a pose-free discriminator built upon EG3D~\cite{Chan2022eg3d}, which embeds a pose predictor in the discriminator to predict the pose of input images in the discriminator for training. It also predicts the distribution of poses in the training datasets used for renderings in the  3D generator.
To generate 3D models with different styles, many researchers explore domain adaptation of trained 3D GANs.
For instance, in 3DAvatarGAN~\cite{abdal20233davatargan}, knowledge from a stylized 2D generator is distilled to a pre-trained 3D generator for domain adaptation. Song~\etal~\cite{song2022diffusion} use pre-trained text-to-image diffusion models to adapt a pre-trained 3D generator to a new text-defined domain, based on the score distillation sampling technique proposed in DreamFusion~\cite{poole2022dreamfusion}. In comparison to these works, our approach focuses on generating calibrated data and efficiently using it to train 3D GANs.

\begin{figure}[t]
	\centering
	\includegraphics[trim=0cm 0cm 0cm 0cm, clip=true,width=1\linewidth]{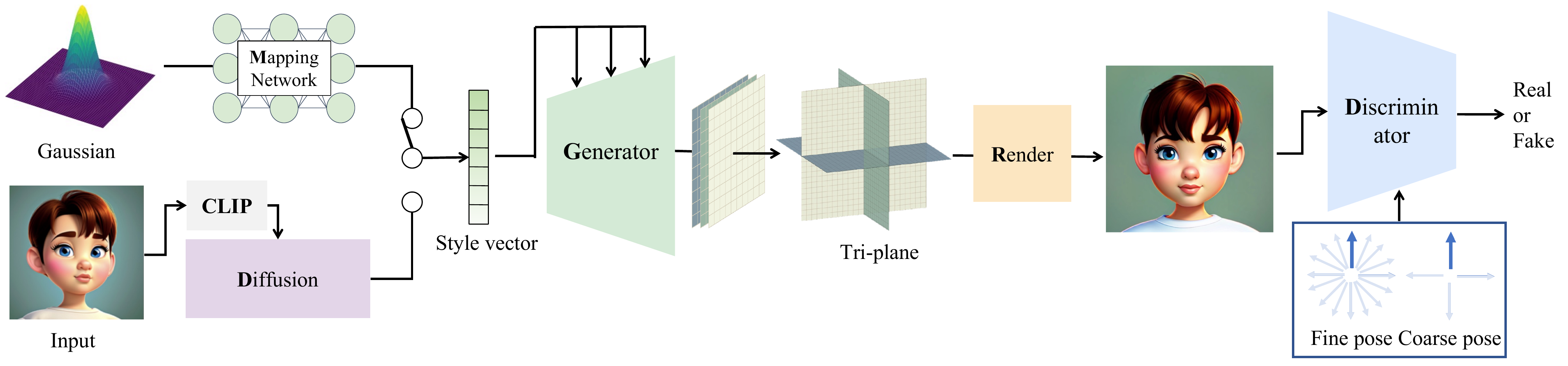}
	\caption{   The overall structure of our networks, which supports unconditional  generation and conditional generation with image inputs. After finishing training of  the unconditional 3D GAN, we train a conditional latent diffusion model to replace the mapping network for conditional avatar generation.  }
	\label{fig:main_net}

\end{figure}

\section{Method}

This section presents the detailed framework of our proposed method for generating stylized 3D avatars trained with multi-view images. We begin by discussing the preparation of multi-view images for training our 3D generator in Section~\ref{sec:image}. Next, we introduce our coarse-to-fine discriminators in Section~\ref{sec:discriminators}, which are designed to address the image-pose misalignment issue in the dataset. Finally, we present our proposed latent diffusion model in the style space that supports image-conditioned 3D generation in Section~\ref{sec:guide}. 
The overview of our framework is illustrated in Fig.~\ref{fig:main_net}.

\subsection{Generating Multi-View Images}
\label{sec:image}

\begin{figure}[t]
	\centering
	\includegraphics[trim=0cm 0cm 0cm 0cm, clip=true,width=1\linewidth]{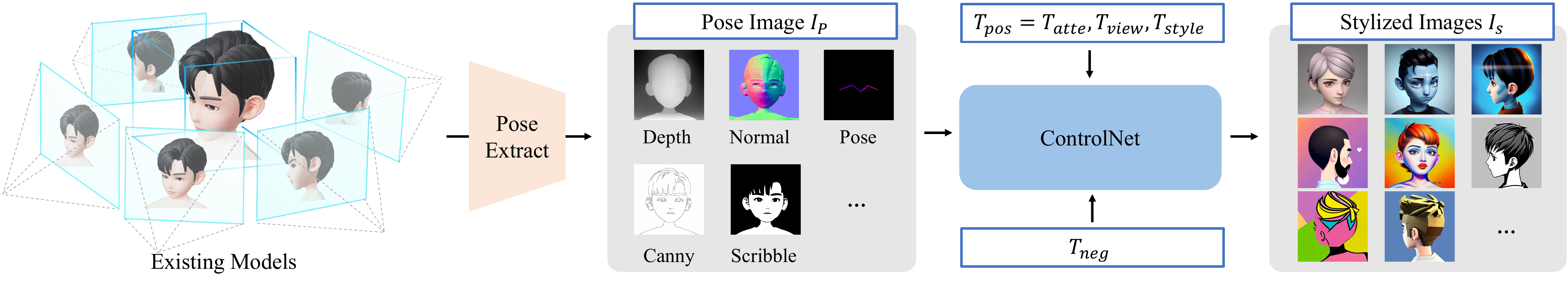}
	\caption{  The pipeline to generate the multi-view image dataset under the guidance of pose images. We extract poses from existing 3D models to guide image synthesis of different views. The styles of avatars are controlled by text prompts. }
	\label{fig:main_data}

\end{figure}

Our framework leverages ControlNet~\cite{zhang2023adding} to produce multi-view images with pose guidance, enabling the definition of avatar styles through textual descriptions.
The pipeline for dataset generation   is illustrated in Fig.~\ref{fig:main_data}.
Specifically,  ControlNet~\cite{zhang2023adding}, represented as $\mathcal{C}_{\theta}$, receives a pose image $I{\text{p}}$ and a text prompt $T$ as input and, in response, generates a stylized image $I_{s}$: $I_{s} = \mathcal{C}_{\theta} (I{\text{p}}, T)$. The text prompt $T$ comprises a positive prompt and a negative prompt: $T= (T_\text{pos}, T_\text{neg})$, which respectively specify the desired and undesired characteristics in the synthesized images.

We use existing 3D avatar models from an engine  to provide pose images for guidance. As a strategic measure to extract pose images from  avatar models,  we designate the center of the avatar's head as the origin of the world coordinate system. This tactic provides a stable and consistent reference point for camera movements. Cameras are then oriented towards this origin and rotated at a predetermined radius to create multi-view images. The yaw and pitch angles of the avatar's front face are assumed to be zero degrees. In accordance with this assumption, the camera position is randomly sampled within a yaw range of -180 degrees to 180 degrees and a pitch range of -30 degrees to 30 degrees.
We investigate three guidance types to generate the pose image $I_{\text{p}}$ : depth maps, human pose (Openpose~\cite{openpose}), and hybrid guidance that synergistically incorporates both depth maps and human pose. 
Both types of pose images are formed as RGB images. As  pose images are generated within engines, we concurrently obtain the camera parameters $c$ of the synthesized images $I_{s}$. This simultaneous acquisition is important as it enables successful training of 3D GANs.

\textbf{View-related prompts.} As the training ControlNet~\cite{zhang2023adding} relies on pseudo pose labels provided by pre-trained human pose estimators, it often fails to synthesize avatars with large face angles,  resulting in avatars with random poses. This issue is also prevalent when using depth guidance, as the depth maps may not accurately reflect the poses, particularly in the backside of the head.
To mitigate such failure cases, we incorporate view-related prompts $T_\text{view}$ into the positive prompt for generating specific views such as ``side view of faces" and ``backside of the head", to improve accuracy.
Additionally, we introduce negative prompts associated with invisible facial features, such as ``eyes" and ``noses", for different views in $T_\text{neg}$. This strategic incorporation significantly diminishes the number of failure cases and improves the generation of multi-view images.

\textbf{Attribute-related prompts.} StableDiffusion~\cite{stable_diffusion} tends to create biased avatars with similar facial attributes, resulting in limited diversity in the generated dataset. 
To counteract this bias, we manually introduce attribute-related prompts $T_\text{att}$ with the objective of enhancing the diversity of created avatars.
 These prompts encompass various aspects including hairstyles, facial expressions, and eye shapes. We incorporate 20 different facial attributes to augment avatar variety. During the generation process, we randomly sample five facial attributes and select one category for each attribute.  Consequently, our positive prompts consist of three parts: style-related prompts $T_\text{style}$, view-related prompts $T_\text{view}$, and attribute-related prompts $T_\text{att}$: $T_\text{pos}= \{T_\text{style}, T_\text{view}, T_\text{att} \}$.  We also demonstrate in our experiments that instead of defining the style of avatars by texts, we can also let the model learn the style from a few examples of images based on LoRA~\cite{hu2021lora}.

\subsection{Addressing the Issue of Image-Pose Misalignment }
\label{sec:discriminators}

In the context of training a 3D generator~\cite{chan2021pi, niemeyer2021giraffe, gu2022stylenerf, Chan2022eg3d} with synthesized multi-view images, a critical challenge that often arises is the problem of image-pose misalignment. 
This problem arises when the pose of the avatar in the generated images does not precisely match the pose image. We observe that the ControlNet~\cite{zhang2023adding} accurately generates front-face images of avatars, likely due to the presence of discernible facial features that facilitate alignment with the provided guidance.
However, the synthesis of side or rear views of avatars presents a considerable challenge, leading to situations where raw pose annotations from the engine no longer fit the images generated.

To mitigate this issue, we introduce a novel coarse-to-fine discriminator specifically designed to support learning with multi-view images, even in scenarios where pose annotations may not be entirely accurate. Each image is associated with two distinct types of pose annotations: a fine pose label $c_\text{fine}$ that corresponds to a more accurate pose annotation, and a coarse pose label $c_\text{coarse}$ that provides a generalized indication of the image's view, as illustrated in Fig~.\ref{fig:main_net}. Given that the camera maintains a fixed radius to render images from avatars, we can simplify the pose representations to yaw and pitch representations.
To generate these pose labels, we divide all rendered views into $N_{\textbf{group}}$ distinct groups based on their yaw and pitch values. Each group is then assigned a unique one-hot yaw representation and a one-hot pitch representation. To derive fine pose labels, we allocate a large group number, and conversely, a small group number is assigned to obtain coarse pose labels. Both types of labels are represented by concatenated one-hot representations of yaw and pitch. The final pose labels used in the discriminator are formed by concatenating the fine labels and coarse labels: $c = c_\text{fine} || c_\text{coarse}$.
During the training process, one type of pose annotation is  sampled for use while the other is set to zero. Views that demonstrate high alignment accuracy, which we refer to as "confident views", are assigned a high sample probability $p_{\text{h}}$ of fine pose annotations. Conversely, views that are less confident and consequently exhibit a lower degree of alignment accuracy are assigned a lower sample probability $p_{\text{l}}$ for fine pose annotations. We define confident views as those views close to the front face, as these are most likely to align accurately with the generated images based on our empirical observations.

\subsection{Image-Guided 3D Generation through Latent Diffusion}
\label{sec:guide}
In EG3D~\cite{Chan2022eg3d}, the authors explore the conditioned face generation process using pivotal tuning~\cite{roich2022pivotal}, a method that optimizes style codes and generators to align the output with a target input image. 
For 3D GANs, the rendered views of the input images, \ie, the poses, should additionally be provided for rendering. However, accurately estimating the poses of stylized avatars can be challenging, particularly for certain complex styles.
To address this challenge, we develop a conditional diffusion model that operates in the latent style space $\mathcal{W}$ of StyleGAN~\cite{karras2019style, karras2020analyzing}. We randomly sample image and style vector pairs from our trained 3D generators to learn the diffusion models. Specifically, we render the front image of a randomly generated 3D avatar using our 3D generators and record its style vector. The diffusion model's objective is to diffuse the style vector from noise, guided by the rendered front image.
We utilize the PriorTransformer~\cite{ramesh2022hierarchical} as our diffusion model $\bm{\epsilon}_\theta$, which receives a noisy style vector $\bm{w}$ and the front image's CLIP-encoded embedding $\bm{y}$ as inputs and predicts the noise $\bm{\epsilon}$. During training, we adopt the approach used in classifier-free diffusion guidance~\cite{ho2022classifier}, where the condition embedding is randomly zeroed with a probability $p_{\text{drop}}$.
During inference, we can adjust the guidance strength $\lambda$ to generate 3D avatars towards our provided condition:
\begin{equation}
  {\bm{\epsilon}}_\theta(\bm{y},\bm{z}) = \lambda \bm{\epsilon}_\theta(\bm{w},\bm{y}) + (1-\lambda) \bm{\epsilon}_\theta(\bm{w}).
\end{equation}
Upon completion of the training process, we can substitute the style mapping network in the original 3D generator with our learned diffusion models to generate 3D avatars conditioned on an input image. This approach eliminates the need to estimate the pose of the input image for rendering, thus improving the accuracy of generating stylized avatars.

\begin{table}[t]

\centering

{%
\begin{tabular}{ll}
\toprule[1.2pt]
\multicolumn{1}{l}{Method}      &FID$\downarrow$ \\ \hline\hline

EG3D~\cite{Chan2022eg3d}   & 7.8       \\  
PoF3D~\cite{shi2023learning}  & 20.9      \\\Xhline{1pt}
\textbf{CoF (Ours)}   & \textbf{5.6}        \\  
\Xhline{1.2pt}
\end{tabular}%
}

\caption{ 
Comparison of our coarse-to-fine (\textbf{CoF}) discriminator with existing methods. Our proposed method outperforms the compared methods with remarkable advantages.}

\label{table:ablation}
\end{table}

\section{Experiments}

In this section, we provide the key results of our experiments. For a more comprehensive analysis and experiment details, we refer the reader to our appendix.

\textbf{Dataset.} 
We collect  50  avatar styles to synthesize data in our experiments.
 All the generated images are in the resolution of $512 \times 512$. 
 For analysis and ablation study, we construct a mix-style dataset with 500,000 images evenly sampled from 50 styles, which adopts a hybrid guidance strategy. This allows us to test our model's ability to generalize across different styles.
When using depth as the pose images, we utilize the Midas model~\cite{Ranftl2022} to extract depth maps from 100,000 avatars created by existing engines. For human pose guidance, we render Openpose~\cite{openpose} annotations of different views based solely on one avatar in the engine. 
To augment our datasets during training, we horizontally flip the synthesized images and pose labels.

\subsection{Results}

\begin{figure}[t]
	\centering
	\includegraphics[trim=0cm 0cm 0cm 0cm, clip=true,width=1\linewidth]{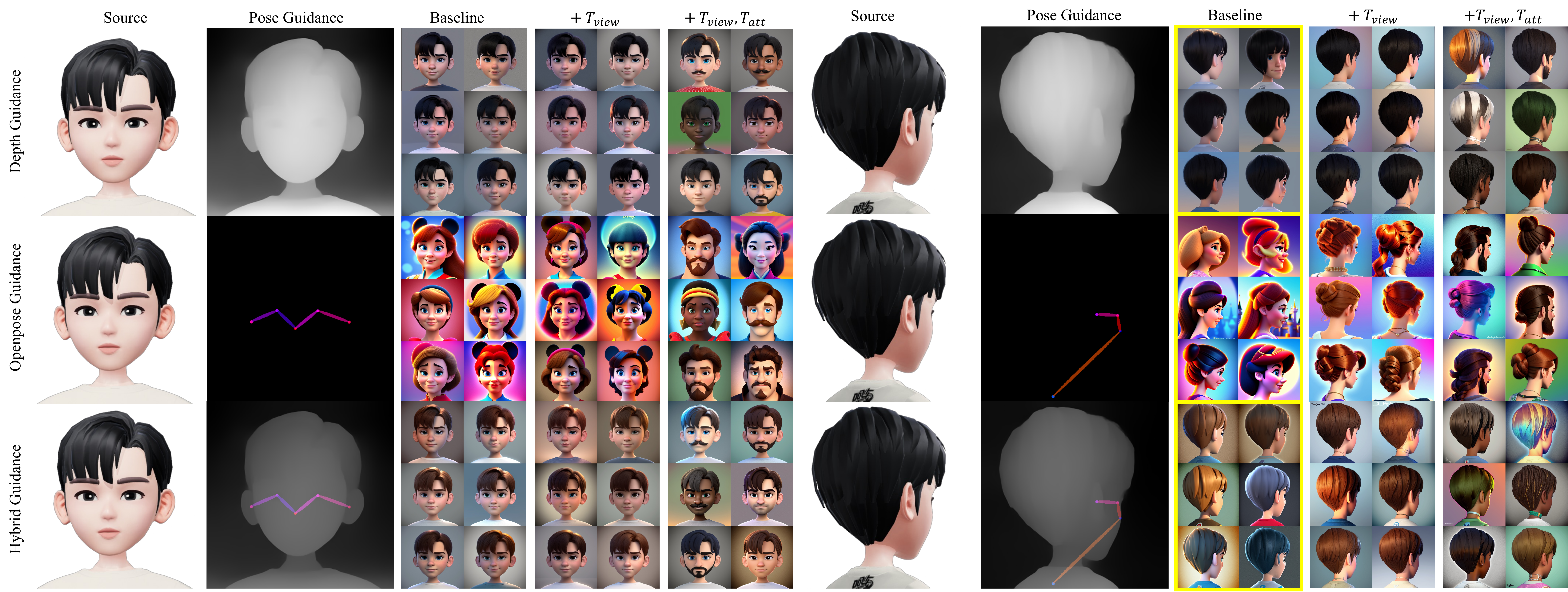}
	\caption{ Comparison of data generated by different guidance strategies and prompts. We can observe how view-specific prompts and attribute-related prompts can effectively enhance generation accuracy and diversity.}
	\label{fig:guidance}

\end{figure}

\textbf{Influence of guidance and prompts on dataset construction.}
In the beginning, we investigate the impact of different types of guidance and prompts on dataset construction. Fig.~\ref{fig:guidance} illustrates the results of our experiments.
When it comes to image generation of the back side of heads, we find that view-specific prompts are particularly effective in eliminating failure cases (see the yellow box).
In addition, we observe that incorporating attribute-related prompts greatly enhances the diversity in the appearance of the generated avatars.
The most effective guidance strategy we find is the hybrid guidance approach, which results in overall better quality and stability of the generated avatars. 
Overall, our findings suggest that careful consideration of guidance and prompts is crucial in constructing high-quality datasets for image generation tasks.

\textbf{Effectiveness of coarse-to-fine discriminators.}
We proceed to validate the effectiveness of our developed coarse-to-fine discriminators, as illustrated in Table~\ref{table:ablation}. We utilize the Fréchet Inception Distance (FID)~\cite{heusel2017gans} as our evaluation metric for a quantitative comparison.
We benchmark our model against the rudimentary discriminator design  in EG3D~\cite{Chan2022eg3d}. Following this, we compare our method with   PoF3D~\cite{shi2023learning}, which incorporates a pose predictor in the discriminator, thereby eliminating the need for pose labels during the training phase.
As evidenced by the results, our coarse-to-fine discriminators significantly outperform the compared methods, demonstrating the superior efficacy of our design. We observe that PoF3D yields subpar results on our generated datasets. A plausible explanation for this could be that predicting the poses of stylized avatars can present substantial challenges due to the high degree of variation in styles and the potential complexity of poses.
The more detailed analysis of our proposed coarse-to-fine discriminator and qualitative comparisons are further elaborated in our appendix.

\begin{figure}[t]
	\centering
	\includegraphics[trim=0cm 0cm 0cm 0cm, clip=true,width=1\linewidth]{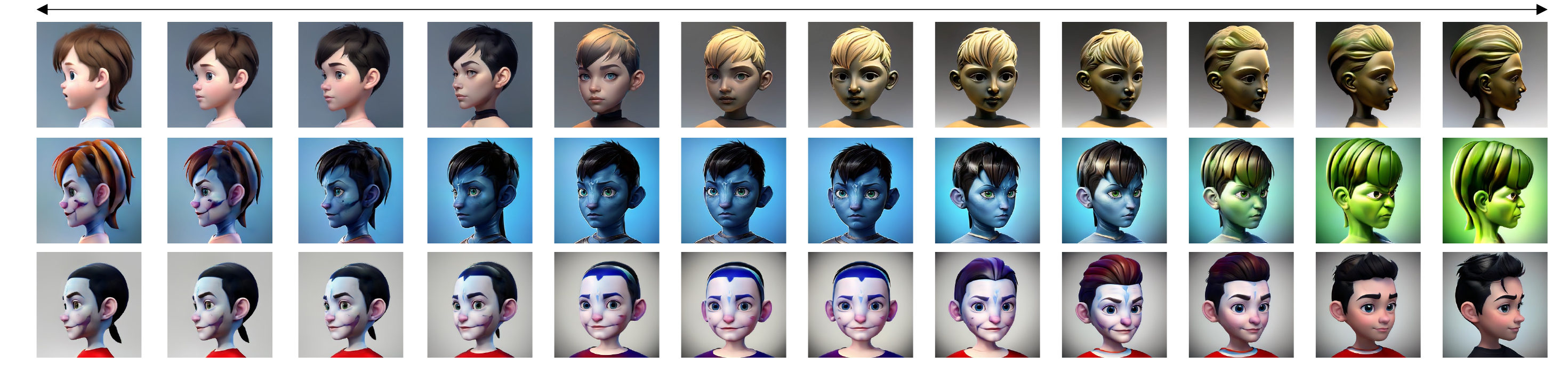}
	\caption{Latent space walk experiment. We linearly shift the input noise and the rendered view from one to the other. The appearance and geometries are changed smoothly. }
	\label{fig:walk}

\end{figure}

\textbf{Latent space walk.}
One intriguing technique to gauge the quality of a learned GAN is to perform a latent space walk. This involves randomly selecting two input vectors and performing a linear interpolation between them. Concurrently, the rendering angle is linearly changed from left to right. This process enables us to observe how the GAN generates images as we navigate through its latent space. As exhibited in Fig.~\ref{fig:walk}, our model is capable of producing visually coherent and diverse images as we traverse its latent space. Moreover, the linear changes in rendering angle allow us to observe how the model responds to variations in viewpoint, which is an important aspect of 3D generation.

\begin{figure}[t]
	\centering
	\includegraphics[trim=0cm 0cm 0cm 0cm, clip=true,width=1\linewidth]{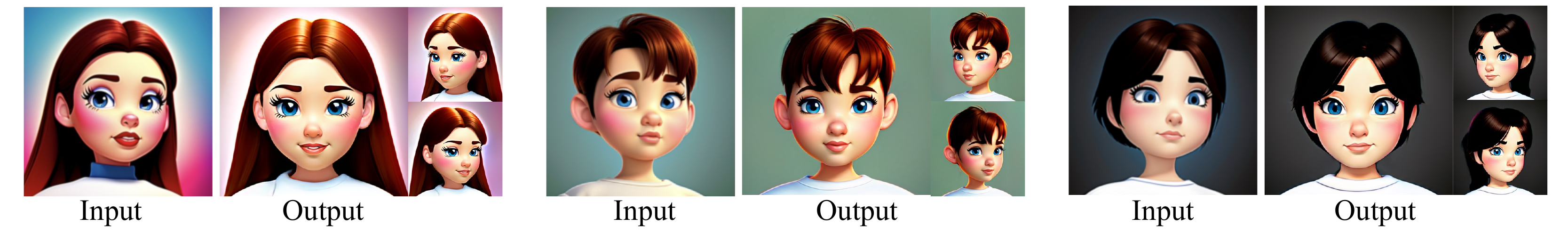}
	\caption{Results of conditional avatar generation with image input. Given the input image on the left, our diffusion predicts the style vector used for generating 3D avatars displayed on the right. }
	\label{fig:condition}

\end{figure}

\textbf{Validation of Image-Conditioned 3D Generation}
To validate the efficacy of our image-conditioned 3D generation approach, we withhold a few images during multi-view image generation as the testing input images. The results of our experiments are shown in Fig.~\ref{fig:condition}.
Our diffusion model in the latent style space is able to effectively generate 3D models under the guidance of the input images. Even though there are minor differences in appearance between the generated avatars and the input images, it is worth noting that our conditional diffusion model effectively captures facial features.

\begin{figure}[t]
	\centering
	\includegraphics[trim=0cm 0cm 0cm 0cm, clip=true,width=1\linewidth]{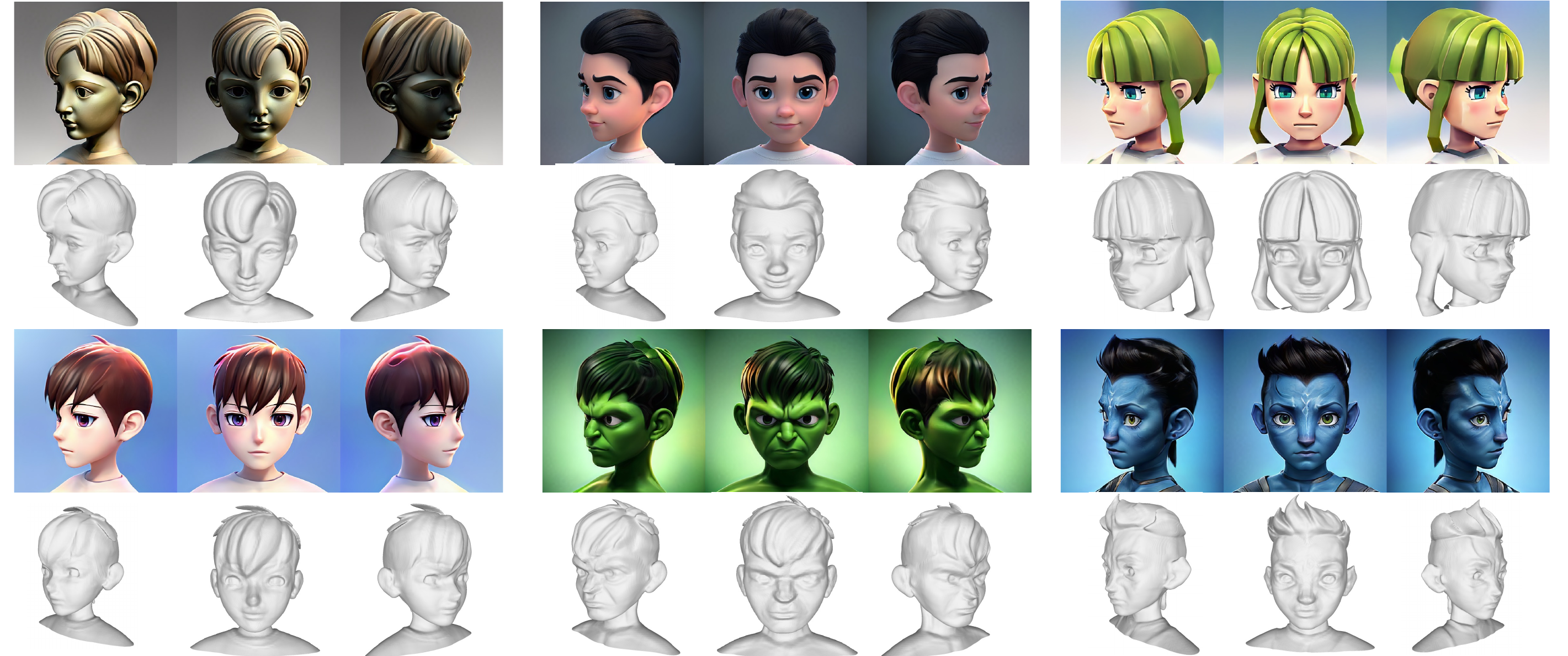}
	\caption{Visualization of exported meshes from our generated 3D avatars. We utilize marching cube algorithms to extract meshes from our learned 3D generation model. Our method is capable of generating avatars with
unique geometries that accurately reflect the desired style. }
	\label{fig:mesh}

\end{figure}

\textbf{Visualization of meshes.} Next, we provide the visualization of meshes extracted from Tri-planes using marching cubes algorithms~\cite{lorensen1987marching}, following EG3D~\cite{Chan2022eg3d}. The resulting meshes and rendered RGB images are presented in Fig.~\ref{fig:mesh}, which portrays the accuracy of our model in generating avatars with realistic geometries. Notably, the geometries of avatars generated by our framework exhibit variance across different styles, suggesting that our approach transcends the mere modification of the appearance of existing 3D avatar models. Instead, our method is capable of generating avatars with unique geometries that accurately reflect the desired style.

\begin{figure}[t]
	\centering
	\includegraphics[trim=0cm 0cm 0cm 0cm, clip=true,width=1\linewidth]{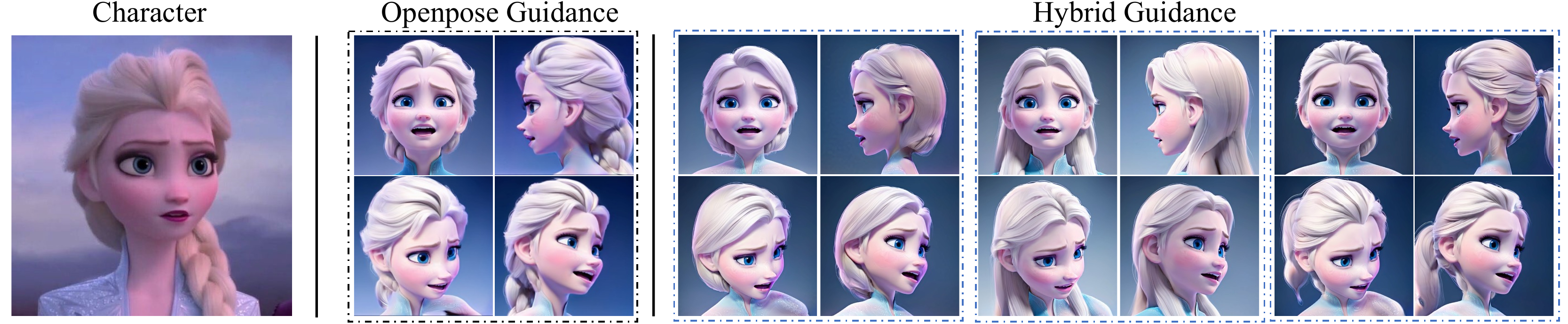}
	\caption{ 3D cartoon character reconstruction using our framework. We used 10 images collected from the Internet to finetune a LoRA + StableDiffusion model for training image generation. }
	\label{fig:lora}

\end{figure}

\textbf{LoRA-based cartoon character reconstruction.}
So far, we have demonstrated the effectiveness of our model, which is trained using data generated by text-defined styles. We next explore the use of a LoRA~\cite{hu2021lora} model based on StableDiffusion~\cite{stable_diffusion}, which can replace the text-defined style with a few example images of a style or subject. To test this approach, we select a well-known cartoon character and gather 10 images of it from the internet to train the LoRA model. Once the training is completed, we generate multi-view images using the learned LoRA model, as outlined in Section~\ref{sec:guide}. We employ Openpose~\cite{openpose} and a hybrid guidance strategy for this experiment.  Fig.~\ref{fig:lora} presents the results of this experiment.
As shown in the figure, when Openpose~\cite{openpose} is used as guidance, our model faithfully reconstructs the 3D model of the character. Conversely, the hybrid guidance strategy enables us to diversify the appearance of the cartoon characters, such as by altering their hairstyles. This result underscores the flexibility of our approach in adapting to different styles and guidance strategies, which could be particularly useful for creating diverse and dynamic avatars or characters.

\section{Conclusion}
In this paper, we introduce a novel framework to  generate stylized 3D avatars by utilizing pre-trained text-to-image diffusion models. The framework offers the ability to define styles and facial attributes using text prompts, which greatly enhances the flexibility of avatar creation. The proposed  coarse-to-fine discriminator can effectively address the issue of misalignment between generated training images and poses, thereby improving the utilization of image data with inaccurate pose annotations. Finally, an additional conditional generation module based on latent diffusion has been developed, enabling the generation of 3D models based on image input.

{\small

\bibliography{reference.bib}{}
\bibliographystyle{plain}
}

\clearpage
\onecolumn
\appendix

\begin{center}
\textbf{\large Appendix}
\end{center}

\section{Introduction}
Our appendix provides additional information about our research. The content is organized as follows:

\begin{itemize}
\item Firstly, in Section~\ref{sec:implementation}, we provide implementation details of our experiments. This section includes more information about dataset generation, network structures, and training.
\item Secondly, in Section~\ref{sec:more_analysis}, we present additional experiment results with analysis. This section includes findings that were not included in the main paper.
\end{itemize}

\section{Implementation Details}
\label{sec:implementation}
\subsection{Dataset Generation}

\textbf{Prompts}.
We first present our detailed prompts for dataset generation.
Table~\ref{tabs:prompt} presents an overview of the prompts we used. For more specific information, Table~\ref{tabs:styles-prompts} and Table~\ref{tabs:att} provide details about the style-related prompts and attribute-related prompts, respectively. The prompts for the first 10 styles were manually designed, while the prompts for the remaining styles were generated using ChatGPT. Similarly, all attribute-related prompts were generated using ChatGPT as well.

\subsection{Training of 3D GANs}
In order to train our 3D GANs effectively, we incorporated both fine and coarse pose annotations. Fine labels were obtained by dividing the yaw range into 40 bins and the pitch range into 15 bins, resulting in one-hot yaw and pitch representations. For coarse labels, we divided the yaw range and pitch range into 3 and 2 bins, respectively. Consequently, the final pose annotation is a 60-dimensional vector representation. In our approach, we define views with yaw values in the range of ${-60^{\circ}, 60^{\circ}}$ and pitch values in the range of ${-15^{\circ}, 15^{\circ}}$ as confident views. During training, we sample fine pose annotations for confident views with a probability of 0.9 ($p_{\text{h}}$), and for the remaining views with a probability of 0.1 ($p_{\text{l}}$).
 
Our 3D GAN is built upon the official implementation of EG3D~\cite{Chan2022eg3d}. Therefore, our training hyperparameters and losses are identical. Specifically, we trained our model on 8 Tesla V100 GPUs with a batch size of 32. The training process consisted of a total of 6000 iterations, which took approximately 5 days. We use an EG3D model pre-trained on FFHQ dataset~\cite{karras2019style} to initialize the parameters of our models, except the layers related to pose encoding. We find such a pre-training strategy can greatly accelerate training. For more details on the training of 3D GANs, please refer to EG3D~\cite{Chan2022eg3d}.
The training of our diffusion model followed a similar setup. It was also trained on 8 Tesla V100 GPUs with a batch size of 32. We used DDPM~\cite{ho2020denoising} with a total of 1000 denoising steps. For inference, we employed DDIM~\cite{song2020denoising} for sampling steps with a total of 50 steps. The diffusion model was trained for 600,000 iterations with a learning rate of 0.0001. Additionally, the probability $p_{\text{drop}}$ of dropping the condition at training time was set to 0.2. During inference, we set the guidance strength $\lambda$ to 5.
During the training of diffusion models, we sampled a 3D avatar from the trained unconditional 3D GANs and rendered its front view as the conditional input image for learning. To augment the training process, we randomly jittered the pose parameters of rendering.

\subsection{Evaluation Metric}
 We utilize the Fréchet Inception Distance (FID)~\cite{heusel2017gans} as our evaluation metric for a quantitative comparison. During the evaluation, we first randomly generate a 3D avatar with our model, then we randomly render a view for evaluation.  It is worth noting that this approach differs slightly from the raw implementations in EG3D\cite{Chan2022eg3d}, where the rendered view for evaluation is always the same as the view provided to the generator for generator pose conditioning (GPC).
The reason for this modification is that we observed a potential bias introduced by GPC. While GPC can enhance the quality of rendering by conditioning on a specific view, it also allows the generator to be aware of the rendered view through this input. Consequently, the rendered view provided through GPC tends to be better than other views of the generated 3D avatars. In order to mitigate this bias during evaluation, we adopt a simple solution by randomly selecting both the generated view and the conditioning view for rendering. This random selection removes the bias and ensures a fair evaluation. For instance, by removing the bias during evaluation, we observed a 0.2 increase in the FID of EG3D~\cite{Chan2022eg3d}.

\section{More Results and Analysis}
\label{sec:more_analysis}
\begin{figure}[t]
	\centering
	\includegraphics[trim=0cm 0cm 0cm 0cm, clip=true,width=1\linewidth]{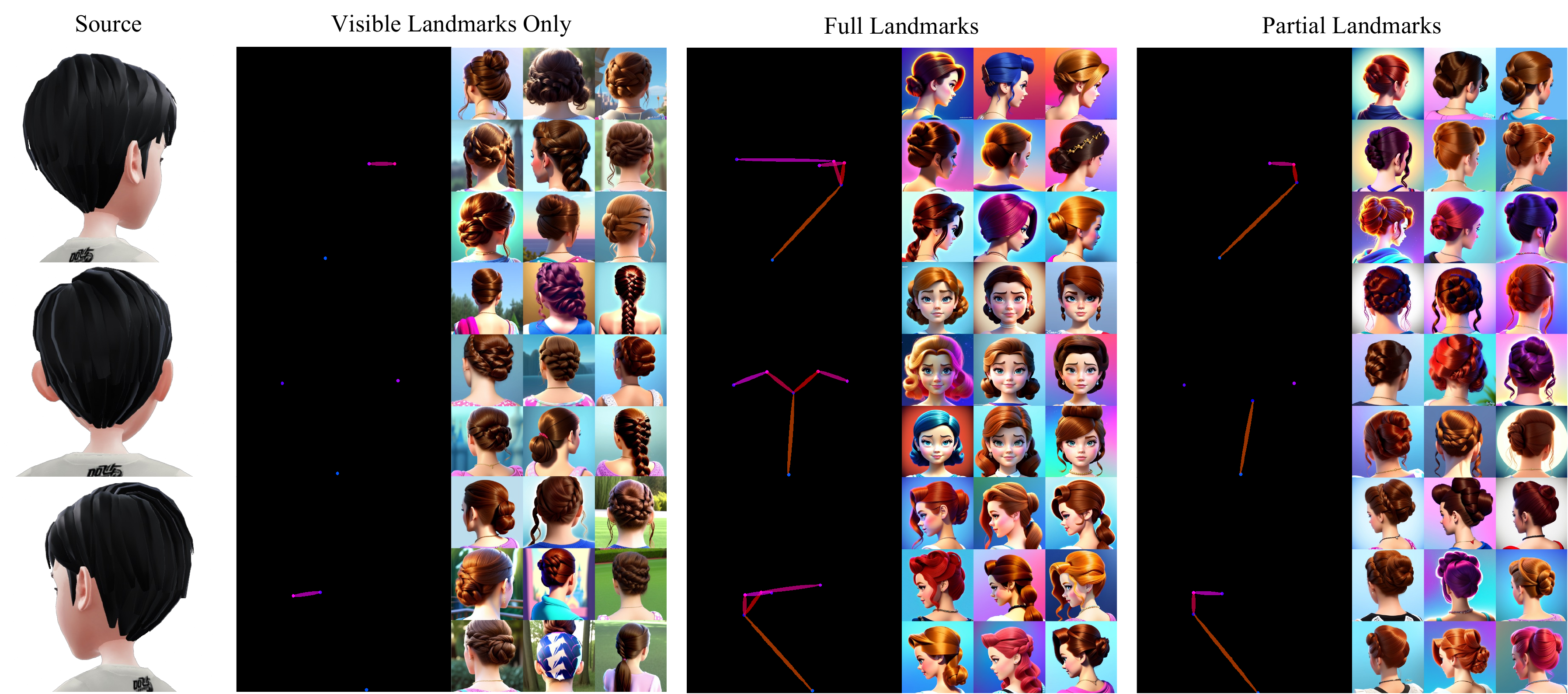}
	\caption{Comparison of strategies for synthesizing Openpose annotations. Synthesizing all facial landmarks and synthesizing visible landmarks alone often result in ambiguity and incorrect poses. Empirically, we found that synthesizing all visible landmarks along with the nose point yields the best results.   }
	\label{fig:landmark}
\end{figure}

\textbf{Openpose synthesis.}
During the synthesis of Openpose annotations from existing 3D models, we explored different strategies, including synthesizing all landmarks or synthesizing visible landmarks only. We compared these strategies in Figure~\ref{fig:landmark}. From the results, we observed that both synthesizing all facial landmarks and synthesizing visible landmarks only often introduced ambiguity and resulted in images with incorrect poses. Through empirical analysis, we found that synthesizing all visible landmarks along with the nose point yielded the best results. However, it is important to note that even with this strategy, the alignment is not perfect, which motivated the design of our coarse-to-fine discriminators.

\begin{figure}[t]
	\centering
	\includegraphics[trim=0cm 0cm 0cm 0cm, clip=true,width=1\linewidth]{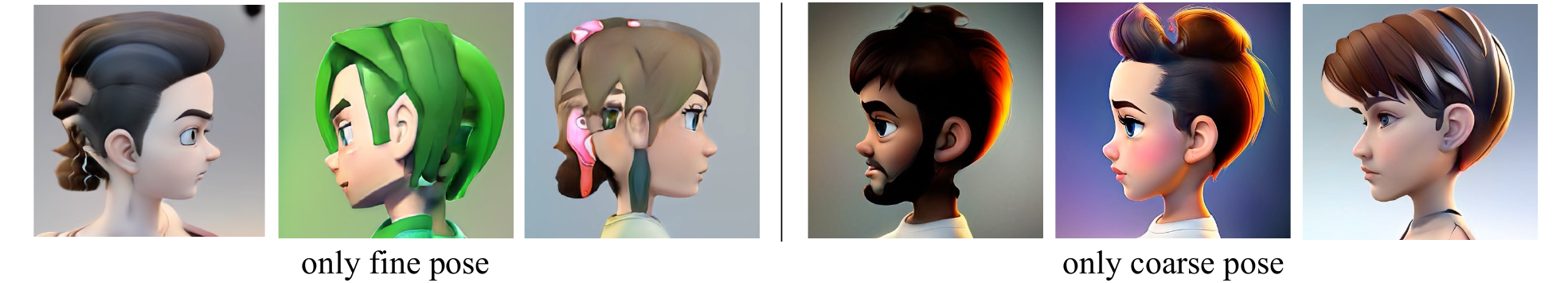}
	\caption{ Typical failure cases in baselines. \textbf{Left:} When image-pose misalignment issues exist in the training images and fine poses are used for training, the generation of the back side of the head is poor due to the over-reliance on pose labels. \textbf{Right:} Using only coarse poses for training may cause the model to converge to distorted 3D avatars that appear unrealistic. In contrast, our model utilizes both coarse and fine pose labels, effectively addressing this problem.   }
	\label{fig:failure}
\end{figure}

\textbf{Ablation study.}
To conduct an ablative analysis, we compared our model with baselines that only utilized fine pose annotations or coarse pose annotations alone, as shown in Table~\ref{tables:ablation}. The comparison revealed that our proposed design outperforms the two baselines, indicating the effectiveness of our coarse-to-fine discriminators. 
We further visualized the typical failure cases of the two baselines in Figure~\ref{fig:failure}. As depicted, when there is a pose-image misalignment issue in the training sets, using fine but inaccurate pose labels for training results in distortion in the back side of the avatar. On the other hand, solely relying on coarse poses for training may cause the model to converge to distorted 3D avatars that appear unrealistic. Our model, by incorporating both fine and coarse pose labels, effectively mitigates these issues and produces more accurate and realistic 3D avatars.

\begin{figure}[t]
	\centering
\includegraphics[trim=0cm 0cm 0cm 0cm, clip=true,width=1\linewidth]{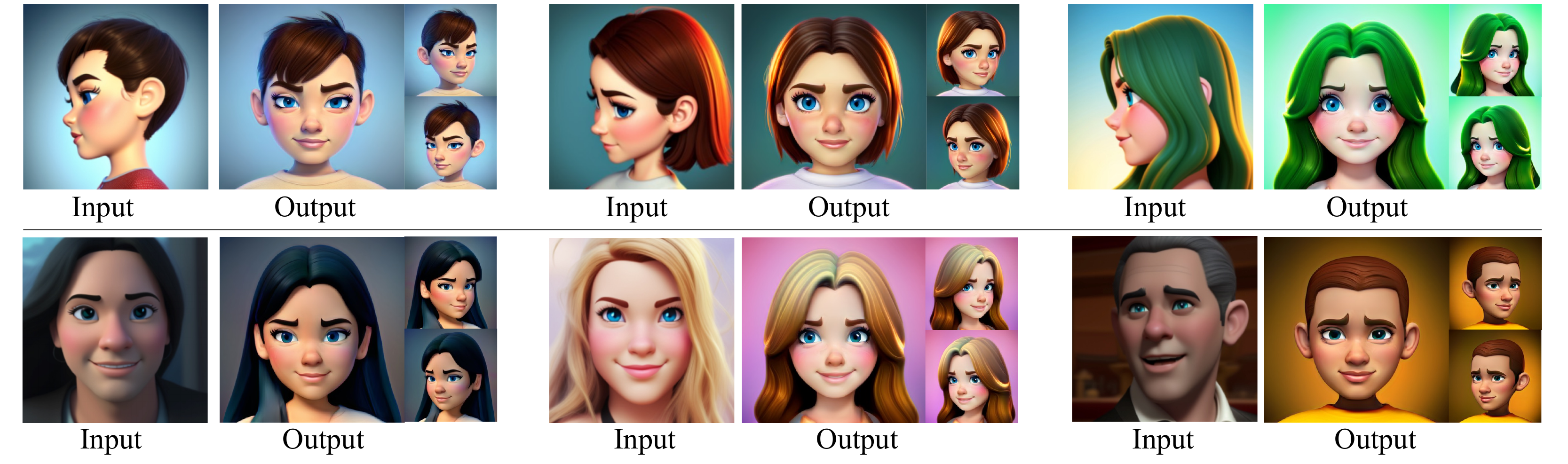}
	\caption{ More results on conditional avatar generation. \textbf{Top:} Input image with large pose angles. \textbf{Down:} Our of domain images as input. To generate out-of-domain images, we utilized StableDiffusion to transform a real face photo into the Disney style, which served as the input to our model. }
	\label{fig:condition2}
\end{figure}

\textbf{Further analysis on conditional avatar generation.} In addition to our main experiments, we conducted an in-depth analysis of two special cases of conditional avatar generation. The first case involves input images with a large face angle away from the front face. The second case explores the generation of avatars from out-of-domain images. To generate out-of-domain images, we utilized StableDiffusion to transform a real face photo into the Disney style, which served as the input to our model. The results of these experiments are presented in Fig.~\ref{fig:condition2}.
From the results, we observe that even though the conditional input during training is a front face image sampled from a trained unconditional 3D GAN, our model is still capable of handling inputs with large pose angles and out-of-distribution images. This finding suggests the effectiveness of our design, which incorporates diffusion in the style space, enabling robust generation of avatars under varying conditions.

\begin{table}[t]

\centering

{%
\begin{tabular}{ll|l}
\toprule[1.2pt]
\multicolumn{1}{l}{Coarse} &\multicolumn{1}{l}{Fine}     &FID$\downarrow$ \\ \hline\hline

 \checkmark&  &   7.1     \\  
& \checkmark&    7.0   \\
\checkmark & \checkmark &  \textbf{5.6}      \\  
\Xhline{1.2pt}
\end{tabular}%
}

\caption{ 
Ablation study on coarse-to-fine discriminators. Utilizing both types of pose annotations in the discriminator helps training and outperforms the baselines with only one type of pose annotations.  }

\label{tables:ablation}
\end{table}

\begin{table}[t]
\centering
\begin{tabular}{|l|l|p{10cm}|}
\hline
\multirow{3}{*}{$T_\text{pos}$} & $T_\text{style}$ & Table~\ref{tab:styles-prompts}  \\ \cline{2-3} 
                   & \multirow{3}{*}{$T_\text{view}$} & \textbf{Front view:} ``\textit{face, head}'' \\ \cline{3-3} 
                   & & \textbf{Side view:} ``\textit{side view of face, side face}'' \\ \cline{3-3} 
                   & & \textbf{Back view:} ``\textit{back of head, back side of the head}'' \\ \cline{2-3} 
                    & $T_\text{att}$ &Table~\ref{tabs:att} \\ \hline
\multirow{2}{*}{$T_\text{neg}$} & $T_\text{view}$ & \textbf{Back view:} ``(((nose, mouse, eyes)))
''  \\ \cline{2-3} 
                   & $T_\text{other}$ & ``\textit{strong light, Bright light, intense light, dazzling light, brilliant light, radiant light, Shade, darkness, silhouette, dimness, obscurity, shadow, blur}'' \\ \hline
\end{tabular}
\caption{Overview of the prompts we used for dataset construction.
\label{tabs:prompt}
}
\end{table}

\begin{table}[h]
\centering
\begin{tabular}{|D|C|D|C|}
\hline
Style & Prompt & Style & Prompt \\
\hline
Disney$^{*}$ &  face, high quality, Disney style, Disney movie, Disney, 3D & Sculpture$^{*}$ & Face, high quality, 3D, Sculpture, statue, Sculptures, stone sculpture, wood sculpture, metal sculpture, ceramic sculpture, glass sculpture, statue, carving, portrait sculpture, 3D effect, Stereoscopy \\
\hline
Dragon Ball$^{*}$ & face, animate, high quality, 3D, Dragon Ball, Dragonball Evolution, Dragon Bowl, Japanese anime, manga, 3D effect, Stereoscopy & Avatar$^{*}$ & face, animate, high quality, Avatar, Avatar style, Avatar movie, movie, James Cameron, blue skin \\
\hline
Pixel Art$^{*}$ & Face, animate, high quality, minecraft style, minecraft, video game, sandbox game, 3D & Anime$^{*}$ & Face, animate, high quality, 3D, Japanese anime, manga, 3D effect, Stereoscopy \\
\hline

Sci-Fi$^{*}$ & face, animate, high quality, Character concept art, Sci-Fi digital painting, trending on ArtStation & Hulk$^{*}$ & head, animate, high quality, 3D, Hulk style, Hulk, Green Giant, movie, 3D effect, Stereoscopy, blur background, blurred background \\

\hline
Joker$^{*}$ & face avatar, face, head, cartoon, animate, high quality, 3D, jocker, Jocker, jocker face, 3D effect, Stereoscopy, blur background, blurred background, cute, lovely, adorable & Robot$^{*}$ & face, animate, high quality, Character concept art, cyber robot with human head, Sci-Fi digital painting, trending on ArtStation\\

\hline
Pop Art & Face, high quality, pop art style, vibrant colors, bold lines, comic book style, 3D effect, Stereoscopy & Graffiti & Face, high quality, graffiti style, bold colors, street art, urban style, 3D effect, Stereoscopy, plain background, Solid color background \\
\hline

Surrealism & Face, high quality, Surrealist style, dreamlike, bizarre, abstract, 3D effect, Stereoscopy & American Comics & Face, animate, high quality, Marvel style, superhero, comic book, 3D effect, Stereoscopy\\
\hline

Cubism & Face, high quality, cubist style, geometric shapes, multiple perspectives, abstract, 3D effect, Stereoscopy & Modernism & Face, high quality, contemporary art style, experimental, unconventional, avant-garde, 3D effect, Stereoscopy\\
\hline

Colorful & Face, high quality, colorful style, bright and bold colors, abstract, 3D effect, Stereoscopy & Sand Painting & Face, high quality, sand painting style, intricate patterns, textured, 3D effect, Stereoscopy\\
\hline

Pokemon & Face, animate, high quality, Pokemon style, anime, video game, 3D effect, Stereoscopy & 
Cyberpunk & Face, animate, high quality, neon lights, futuristic, dystopian, cybernetic implants, 3D effect, Stereoscopy\\
\hline

Realistic & Face, high quality, realistic style, lifelike, detailed, 3D effect, Stereoscopy & Cartoon & Face, animate, high quality, cartoon style, exaggerated features, bright colors, 3D effect, Stereoscopy \\
\hline

Steampunk & Face, high quality, steampunk style, Victorian era, gears, clockwork, 3D effect, Stereoscopy & Manga & Face, animate, high quality, manga style, Japanese comics, black and white, 3D effect, Stereoscopy  \\
\hline

Art Nouveau & Face, high quality, Art Nouveau style, flowing lines, organic shapes, 3D effect, Stereoscopy & Expressionism & Face, high quality, Expressionist style, distorted features, bold colors, 3D effect, Stereoscopy\\
\hline

\end{tabular}
\caption{Style-related prompts. The prompts of  the top 10 styles ($^{*}$) are manually designed, while the rest is generated by ChatGPT.  }
\label{tab:styles-prompts}
\end{table}

\begin{table}[h]
\ContinuedFloat
\centering
\begin{tabular}{|D|C|D|C|}
\hline
Style & Prompt & Style & Prompt \\
\hline
Fauvism & Face, high quality, Fauvist style, bold colors, simplified forms, 3D effect, Stereoscopy & Art Deco & Face, high quality, Art Deco style, geometric shapes, metallic accents, 3D effect, Stereoscopy \\
\hline

Pop Surrealism & Face, high quality, Pop Surrealist style, surreal imagery, bright colors, 3D effect, Stereoscopy & Baroque & Face, high quality, Baroque style, ornate details, dramatic lighting, 3D effect, Stereoscopy \\
\hline

Rococo & Face, high quality, Rococo style, pastel colors, ornate details, 3D effect, Stereoscopy & Neo-Expressionism & Face, high quality, Neo-Expressionist style, bold colors, thick brushstrokes, 3D effect, Stereoscopy \\
\hline

Art Brut & Face, high quality, Art Brut style, raw and unrefined, childlike, 3D effect, Stereoscopy & Surrealist Photography & Face, high quality, surrealist photography style, dreamlike, bizarre, abstract, 3D effect, Stereoscopy \\
\hline

Concept Art & Face, high quality, concept art style, imaginative, futuristic, 3D effect, Stereoscopy & Low Poly & Face, animate, high quality, low poly style, geometric shapes, bright colors, 3D effect, Stereoscopy\\
\hline

Art Synthetique & Face, high quality, Art Synthetique style, digital art, abstract, 3D effect, Stereoscopy & Art Informel & Face, high quality, Art Informel style, spontaneous, abstract, 3D effect, Stereoscopy \\
\hline

Tonalism & Face, high quality, Tonalist style, muted colors, atmospheric, 3D effect, Stereoscopy & Chibi & Face, animate, high quality, chibi style, small and cute, exaggerated features, 3D effect, Stereoscopy \\
\hline

Claymation & Face, animate, high quality, claymation style, stop-motion animation, tactile, 3D effect, Stereoscopy & Cutout & Face, animate, high quality, cutout style, paper cutouts, stop-motion animation, 3D effect, Stereoscopy \\

\hline

Flash Animation & Face, animate, high quality, flash animation style, vector graphics, smooth animation, 3D effect, Stereoscopy & Hand-Drawn & Face, animate, high quality, hand-drawn style, traditional animation, pencil and paper, 3D effect, Stereoscopy\\
\hline

Motion Graphics & Face, animate, high quality, motion graphics style, typography, kinetic typography, 3D effect, Stereoscopy & Pencil Test & Face, animate, high quality, pencil test style, rough animation, sketchy, 3D effect, Stereoscopy \\
\hline

Photorealistic & Face, animate, high quality, photorealistic style, lifelike, detailed, 3D effect, Stereoscopy & Rubber Hose & Face, animate, high quality, rubber hose style, 1920s animation style, flexible limbs, 3D effect, Stereoscopy \\
\hline
Sand Animation & Face, animate, high quality, sand animation style, sand art, stop-motion animation, 3D effect, Stereoscopy & Stop-Motion & Face, animate, high quality, stop-motion animation style, puppetry, claymation, 3D effect, Stereoscopy \\
\hline

Traditional & Face, animate, high quality, traditional animation style, hand-drawn, frame-by-frame, 3D effect, Stereoscopy & Vector & Face, animate, high quality, vector style, clean lines, scalable, 3D effect, Stereoscopy \\
\hline

Woodcut & Face, animate, high quality, woodcut style, carved lines, black and white, 3D effect, Stereoscopy & Zoetrope & Face, animate, high quality, zoetrope style, pre-cinema animation, spinning cylinder, 3D effect, Stereoscopy \\
\hline

\end{tabular}
\caption{Style-related prompts. (Continued) The prompts of  the top 10 styles ($^{*}$) are manually designed, while the rest is generated by ChatGPT.}
\label{tabs:styles-prompts}
\end{table}

\begin{table}[ht]
\centering
\small
\begin{tabular}{|D|E|}
\hline
\textbf{Attribute} & \textbf{Description} \\
\hline
Eye Shape & small-eyed, big-eyed, almond-shaped eyes, round eyes, narrow eyes, deep-set eyes, protruding eyes, close-set eyes, wide-set eyes \\
\hline
Eyebrows & thick-browed, sparse-browed, arched eyebrows, straight eyebrows, bushy eyebrows, thin eyebrows, unibrow \\
\hline
Eyelashes & long-lashed, short-lashed, thick lashes, sparse lashes, curled lashes \\
\hline
Cheeks & rosy-cheeked, pale-cheeked, chubby cheeks, hollow cheeks, high cheekbones, low cheekbones \\
\hline
Ears & big-eared, small-eared, attached earlobes, detached earlobes, ear piercings \\
\hline
Expression & happy, sad, angry, surprised, tired, anxious, nervous, handsome, ugly, smiling, frowning, scowling, smirking, pouting, grinning, winking, raising eyebrows \\
\hline
Facial Hair & moustache, beard, goatee, stubble, clean-shaven, sideburns \\
\hline
Eye Color & blue eyes, black eyes, brown eyes, green eyes, hazel eyes, gray eyes \\
\hline
Skin & freckle, mole, wrinkled, smooth skin, acne-prone skin, oily skin, dry skin, sensitive skin \\
\hline
Race & Asian, European, Africans, Latino, Middle Eastern, Indian, mixed race \\
\hline
Age & old, young, middle-aged, elderly, baby-faced, mature \\
\hline
Chin & thick-lipped, thin-lipped, cleft chin, dimpled chin, pointed chin, square chin, round chin \\
\hline
Face Shape & square-faced, thin-faced, round-faced, chubby-faced, pointy-chinned, prominent-chinned, heart-shaped face, oval face, diamond-shaped face \\
\hline
Nose & short-nosed, long-nosed, high-nosed, low-nosed, high-bridged nose, low-bridged nose, upturned nose, downturned nose, button nose, Roman nose \\
\hline
Lips & full-lipped, thin-lipped, downturned lips, upturned lips, bow-shaped lips, heart-shaped lips, thin upper lip, full lower lip \\
\hline
Forehead & high forehead, low forehead, receding hairline, widow's peak \\
\hline
Eye Sockets & deep-set eyes, hooded eyes, almond-shaped eyes, protruding eyes, round eyes, sunken eyes \\
\hline
Facial Features & dimpled chin, cleft chin, birthmark, scar, tattoo, beauty mark, mole, freckles \\
\hline
Facial Contour & sharp jawline, soft jawline, high cheekbones, low cheekbones, narrow face, wide face \\
\hline
Facial Impression & friendly, serious, confident, approachable, intimidating, warm, cold, inviting, unapproachable \\
\hline
Hairstyle & bald, short hair, long hair, curly hair, straight hair, wavy hair, bangs, ponytail, bun, braids, cornrows \\
\hline
\end{tabular}
\caption{Attribute-related prompts. All prompts are generated by ChatGPT.}
\label{tabs:att}
\end{table}



\end{document}